\pdfoutput=1

\documentclass[11pt]{article}

\usepackage[final]{acl}

\usepackage{times}
\usepackage{latexsym}
\usepackage{tcolorbox}
\usepackage{url}
\usepackage{listings}
\usepackage{xparse}
\lstdefinestyle{pythonstyle}{
  language=Python,
  basicstyle=\ttfamily\small,
  keywordstyle=\bfseries\color{blue},
  commentstyle=\itshape\color{green!60!black},
  stringstyle=\color{red!70!black},
  showstringspaces=false,
  breaklines=true,
  frame=single
}

\lstdefinestyle{txtfile}{
    basicstyle=\scriptsize\ttfamily,
    backgroundcolor=\color{white},
    frame=single,
    rulecolor=\color{black!30},
    stringstyle=\color{red},
    numbers=left,
    numberstyle=\tiny\color{gray!50},
    numbersep=5pt,
    breaklines=true,
    breakatwhitespace=true,
    showstringspaces=false,
    tabsize=4,
    captionpos=b
}

\usepackage{tabularray}
\usepackage{color}
\usepackage{booktabs}
\usepackage{siunitx}
\usepackage{graphicx}
\usepackage{array}

\usepackage{listings}
\usepackage{xcolor}
\usepackage{algorithm}
\usepackage{algorithmic}
\usepackage{wrapfig}
\usepackage{tabularx}
\usepackage{subcaption}
\usepackage{pifont}

\usepackage[T1]{fontenc}

\usepackage[utf8]{inputenc}
\usepackage{amsmath}

\usepackage{microtype}

\usepackage{inconsolata}

\usepackage{graphicx}

\usepackage{amsfonts}
\usepackage{booktabs}
\usepackage{multirow}

\title{I-MCTS: Enhancing Agentic AutoML via Introspective Monte Carlo Tree Search}

\author{
 \textbf{Zujie Liang\textsuperscript{1}}\quad
 \textbf{Feng Wei\textsuperscript{1}}\quad
 \textbf{Wujiang Xu\textsuperscript{2}}\quad
 \textbf{Lin Chen\textsuperscript{1}}\quad
 \textbf{Yuxi Qian\textsuperscript{1}}\quad
 \textbf{Xinhui Wu\textsuperscript{1}}
\\
\\
 \textsuperscript{1}Ant Group\;\;\;\;\;\;    
 \textsuperscript{2}Rutgers University
}

\newcommand{\methodname}{\textsc{I-MCTS}}

\begin{document}
\maketitle

\begin{abstract}
Recent advancements in large language models (LLMs) have shown remarkable potential in automating machine learning tasks. 
However, existing LLM-based agents often struggle with low-diversity and suboptimal code generation. 
While recent work~\cite{chi2024sela} has introduced Monte Carlo Tree Search (MCTS) to address these issues, limitations persist in the quality and diversity of thoughts generated, as well as in the scalar value feedback mechanisms used for node selection. 
In this study, we introduce Introspective Monte Carlo Tree Search (\textbf{\textit{I-MCTS}}), a novel approach that iteratively expands tree nodes through an introspective process that meticulously analyzes solutions and results from parent and sibling nodes. 
This facilitates a continuous refinement of the node in the search tree.
Furthermore, we integrate a Large Language Model (LLM)-based value model to facilitate direct evaluation of each node's solution prior to conducting comprehensive computational rollouts. 
A hybrid rewarding mechanism is implemented to seamlessly transition the Q-value from LLM-estimated scores to actual performance scores. 
This allows higher-quality nodes to be traversed earlier.
Applied to the various ML tasks, our approach demonstrates a
4\% absolute improvement in performance compared to the strong open-source AutoML agents, showcasing its effectiveness in enhancing agentic AutoML systems\footnote{Resource available at \url{https://github.com/jokieleung/I-MCTS}}.
\end{abstract}

\section{Introduction}
\label{sec:intro}

Recent advances in large language models (LLMs) have opened new frontiers in automating machine learning (AutoML) tasks~\cite{huang2024mlagentbench,chan2024mle}.
Compared to the traditional AutoML frameworks~\cite{feurer2015efficient, tang2024autogluon}, the LLM-based agents~\cite{sun2024survey} emerge as promising direction 
due to their remarkable capabilities on code generation~\cite{jimenez2023swe}, neural architecture design~\cite{zheng2023can} and model training~\cite{chi2024sela}. 
Overall, these LLM agent-based AutoML systems~\cite{Schmidt_Wu_Jiang,hong2024data,li2024autokaggle,chi2024sela} typically input with a natural language description on the dataset and the problem, 
after which the system directly generates a solution in an end-to-end manner. 
While recent works by \citet{Schmidt_Wu_Jiang} and \citet{hong2024data} have made significant strides in automating the machine learning workflow, replicating the adaptive and strategic behavior of expert human engineers remains a significant challenge. 
The primary limitation lies in their search processes, which typically involve only a single pass or trial. 
This constraint significantly hinders the generation of diverse and highly optimized workflows, a capability that human experts excel at through iterative refinement and strategic decision-making.
Recent work by ~\citet{chi2024sela} introduced Tree-Search Enhanced LLM Agents (SELA), leveraging Monte Carlo Tree Search (MCTS) to expand the search space. 
However, the search space is constrained by its reliance on a pre-generated and fixed insight pool derived from the initial problem description and dataset information.
This static nature inherently limits the diversity and adaptability of the search tree.
Moreover, SELA encounters significant limitations in improving the overall quality of solutions when relying exclusively on scalar experimental performance feedback. 
This reliance on simplistic scalar feedback mechanisms impedes the efficient identification and navigation of optimal pathways in complex machine learning tasks, where flexible re-assessment and adaptive strategies are usually needed.

In this paper, we present \textbf{I}ntrospective \textbf{M}onte \textbf{C}arlo \textbf{T}ree \textbf{S}earch (\textbf{\textit{I-MCTS}}), a novel inference-time scaling paradigm for Agentic AutoML that successfully combines the internal test-time compute (reflection) and external test-time compute (tree search) into one framework.
There are two key innovations in our method. 
First, our \textit{introspective node expansion} dynamically generates high-quality thought nodes through explicit analysis of parent and sibling node states. 
By incorporating reflective reasoning and feedback about prior solutions and their outcomes, this approach enables continuous quality improvement of the search tree nodes. 
Second, we develop a \textit{hybrid rewarding mechanism} that combines: 1) LLM-estimated evaluations predicting node potential through a comprehensive machine learning evaluation criteria, 
and 
2) actual performance scores on the development set from computational rollouts. 
Our adaptive reward blending strategy smoothly transitions Q-value from LLM-estimated values to actual values across search iterations.
This facilitates higher-quality nodes to be traversed earlier. 
Overall, the principal contributions of this work are twofold:

\begin{itemize}
\item We introduce I-MCTS, an innovative approach for agentic AutoML that incorporates an introspective node expansion process and a hybrid reward mechanism. This enhances both the quality and efficiency of tree searches in AutoML workflows.
    
\item Extensive experiments across diverse ML tasks demonstrate our method's superiority, achieving 4\% absolute performance gains over state-of-the-art baselines while maintaining computational efficiency.
\end{itemize}

\section{Related work}
\label{app:related_work}

Recent advances in autonomous language agents have focused on iterative self-improvement and multi-agent coordination. \citet{DBLP:conf/nips/MadaanTGHGW0DPY23} proposed SELF-REFINE, which iteratively refines LLM outputs through self-generated feedback without additional training. Building on iterative refinement, \citet{DBLP:conf/nips/ShinnCGNY23} introduced Reflexion, which leverages natural language feedback and episodic memory to enable LLM agents to learn from errors. \citet{DBLP:conf/icml/ZhouYSWW24} unified reasoning and action through Language Agent Tree Search (LATS).
This is the most related work that also 
introduces self-reflection techniques into the Monte Carlo Tree Search process.
However, the reflection is generated and used after the trajectory fails as additional context for future trials. 
By contrast, our introspective node expansion dynamically generates high-quality thought nodes through explicit analysis of parent and sibling node states, which enables continuous quality improvement of the search tree nodes. 

At the intersection of multi-agent systems, Fleet of Agents~\cite{DBLP:journals/corr/abs-2405-06691} employs genetic particle filtering to coordinate LLM agents, outperforming Tree-of-Thoughts~\cite{yao2023tree} methods in efficiency and accuracy for puzzle-solving tasks. \citet{DBLP:conf/nips/QuZGK24} extended these concepts through recursive introspection, enabling agents to systematically self-improve via structured reflection cycles.


\section{Introspective Monte Carlo Tree Search (I-MCTS)}
\label{sec:method}

\begin{figure}[!t]
    \centering
    \includegraphics[width=0.95\linewidth]{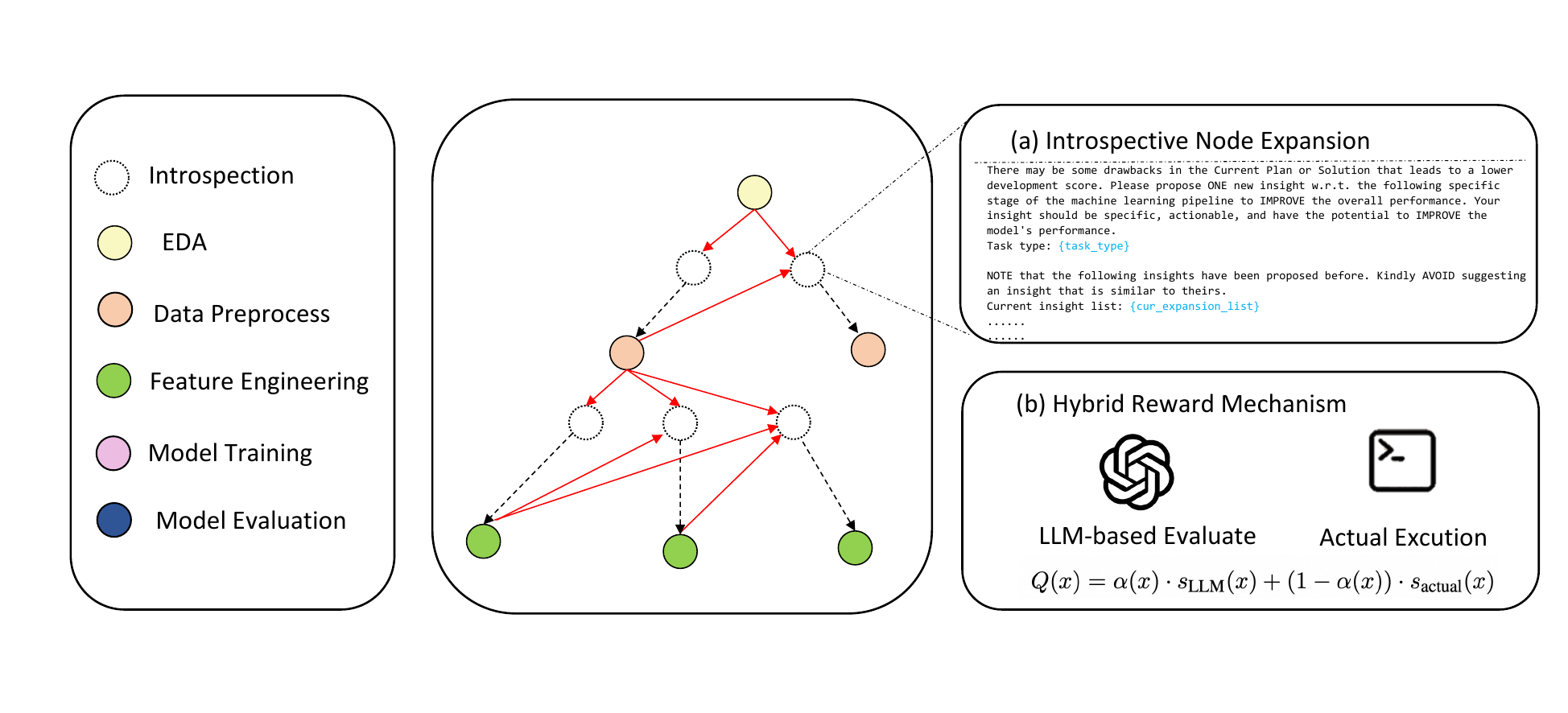}
    \caption{I-MCTS architecture featuring (a) Introspective node expansion through parent/sibling analysis, (b) Hybrid reward calculation combining LLM predictions and empirical scores. The red arrows indicate the introspective feedback loop that continuously improves node quality. For the full prompt design, please refer to Appendix~\ref{app:prompt}.
    }
    \label{fig:pipline}
\end{figure}

As illustrated in Figure~\ref{fig:pipline}, our approach consists of two key components: a search module I-MCTS, and an LLM agent as the experiment executor. 
The Introspective Monte Carlo Tree Search (I-MCTS), builds upon the foundation of traditional MCTS but introduces two key innovations: 
(1) introspective node expansion through reflective solution generation, and (2) a hybrid reward mechanism combining LLM-estimated evaluations with empirical performance scores. 
These components work in tandem to enhance the quality and diversity of thought nodes while improving the efficiency of the search process. 
During each cycle, the selected path is passed to the LLM agent, which plans, codes, executes, and introspects the experiment, providing both scalar and verbal feedback to refine future searches. 
This iterative process continues until the termination criterion is met. 

\subsection{Preliminary}
We formalize the automated machine learning (AutoML) problem as a search process over possible experimental configurations. 
Given a problem description $p$ and dataset $D$ with metadata $d$, the objective is to identify an optimal pipeline configuration $c^* \in \mathcal{C}$ that maximizes a performance metric $s$ on development data. 
The search space $\mathcal{C}$ consists of configurations combining insights $\lambda_i^\tau$ across $T$ predefined stages of the ML workflow: $\tau_1$ (Exploratory Data Analysis), $\tau_2$ (Data Preprocessing), $\tau_3$ (Feature Engineering), $\tau_4$ (Model Training), and $\tau_5$ (Model Evaluation).

The LLM agent $E$ conducts each trial by building practical experimental pipelines from natural language requirements. 
Given an experiment configuration $c$, the agent produces a detailed plan $E_{\text{plan}}$ that consists of a sequence of task instructions $I^{\tau \in T}$ corresponding to each stage of the machine learning process. Next, the agent writes and executes code $\sigma^\tau$ for each task $\tau$ based on the respective instruction and gets the final execution score $s$. 
The complete set of code outputs $\sigma^{\tau \in T}$ is concatenated into a full solution $\sigma_{sol}$ to address the problem. This phase is referred to as $E_{\text{code \& execute}}$.
\begin{align}
    E_{\text{plan}}(p, d, c, LLM) & \rightarrow I^{\tau \in T} \\
    E_{\text{code \& execute}}(I^{\tau \in T}, D, LLM) & \rightarrow (\sigma^{\tau \in T}, s)
\end{align}

\subsection{Introspective Node Expansion}
\label{ssec:introspective}

Unlike static insight pools in prior work~\cite{chi2024sela}, 
I-MCTS utilizes the introspective expansion mechanism that dynamically generates high-quality thought nodes by leveraging solutions and results from parent and sibling nodes. 
This design mirrors the way human expert engineers backtrack and refine their solution dynamically, ensuring the agent can iteratively
improve its decision-making process based on past experiences.
Given a parent node $x_{\text{parent}}$ and its existing sibling nodes $x_{\text{sibling}}$, the introspective expansion process generates a new node $x_{\text{child}}$ by introspecting the solutions and results of these previous nodes. 
Specifically, the introspection mechanism use an LLM to evaluate the solutions $\sigma_{\text{sol}}(x_{\text{parent}})$ and $\sigma_{\text{sol}}(x_{\text{sibling}})$ associated with the parent and sibling nodes and identify strengths, weaknesses, issues and potential improvements in the solutions.
Then, the LLM generates a fine-grained and customized insight $\lambda_{\text{new}}$ that addresses the identified issues or builds upon the strengths. 
This insight is tailored to the current stage $\tau$ of the machine learning pipeline. The new insight $\lambda_{\text{new}}$ is used to create a child node $x_{\text{child}}$, which inherits attributes from the parent node while incorporating the refined insight. 
The child node is then added to the search tree.
This introspection process ensures that each new node is dynamically generated through a thoughtful analysis of prior solutions, leading to a continuous improvement in the quality of the search tree.
Also, the verbalized introspection feedback addresses the solely scalar feedback limitation.

\subsection{Hybrid Reward Mechanism}
\label{ssec:hybrid}

The primary objective of the evaluation phase is to assess the reward for the current node.
For machine learning tasks, a conventional approach~\cite{Schmidt_Wu_Jiang,chi2024sela} involves utilizing the performance metric on the development set as the reward. 
To derive a performance score of the given node involves rolling out from an intermediate state to a terminal state, which is quite costly since processes such as model training cost significant computational time. 
This limits the efficiency of node exploration.
To address these challenges, we introduce a hybrid reward mechanism for node evaluation which combines LLM-estimated evaluations with actual performance scores to guide the search process more efficiently. 
For each node $x$, we employ an LLM-based value model $M_{\text{value}}$ to predict its potential performance. The value model takes as input the solution $\sigma_{\text{sol}}(x)$ and outputs an estimated score $s_{\text{LLM}}(x)$. This score is based on a comprehensive set of machine learning evaluation criteria, which is detailed in Appendix~\ref{app:llm_evaluation_schema}.
The LLM-estimated evaluation allows us to assess the quality of a node's solution before conducting computationally expensive rollouts. This early evaluation helps prioritize nodes with higher potential, improving the efficiency of the search process.
After the "simulation" of a node $x$, we obtain the actual performance score $s_{\text{actual}}(x)$ based on the development set, which reflects the true performance of the node's solution.
To smoothly transition from LLM-estimated values to actual performance scores, we implement an adaptive reward blending strategy. The Q-value $Q(x)$ for a node $x$ is updated as follows:

\begin{align}
    Q(x) = \alpha(x) \cdot s_{\text{LLM}}(x) + (1 - \alpha(x)) \cdot s_{\text{actual}}(x)
\end{align}

where $\alpha(x) = \frac{\gamma}{n_{\text{visits}(x)}+\gamma}$, a blending factor that decreases over the visit count increase, where $\gamma$ is a controlling constant.
This adaptive blending ensures that higher-quality nodes are traversed earlier in the search process, while still incorporating the true performance feedback as it becomes available.

\subsection{Tree Search Process}
\label{ssec:adaptive}

The overall search process can be summarized as:

\paragraph{Selection} At each iteration, we select node according to the UCT formula as:

\begin{align}
    \text{UCT}(x) = \frac{Q(x)}{n(x)} + \alpha_{\text{explore}} \sqrt{\frac{\ln n_{\text{visits}}(x_{\text{parent}})}{n(x)}}
\end{align}

\paragraph{Expansion} Child nodes are generated through the introspective process described in Section~\ref{ssec:introspective}, creating a dynamic search space that evolves based on simulation feedback.

\paragraph{Simulation} Each rollout executes the full pipeline while caching intermediate results. The LLM value model provides real-time estimates $s_{\text{LLM}}$ before computational execution.

\paragraph{Backpropagation} 
Upon simulation completion, the Q value is propagated backwards through the tree structure. 
This process updates the value and visit count of each parent node, from the simulated node to the root. Consequently, nodes associated with superior solutions receive higher priority in subsequent iterations.
This allows nodes representing more promising solutions to be prioritized in future rollouts.
Similar to \citet{chi2024sela}, I-MCTS implements a state-saving mechanism that caches the stage code for each node, which allows it to reuse previously generated code when a new configuration shares components with existing ones.
For the detailed algorithm flow, please refer to Appendix ~\ref{alg:i-mcts}.

\section{Experiment}

\subsection{Experimental Details}
For fair comparison, we follow \citet{chi2024sela} to benchmark our method on 20 machine learning datasets, including 13 classification tasks and 7 regression tasks from the AutoML Benchmark (AMLB) ~\cite{gijsbers2024amlb} and Kaggle Competitions.
Five baselines are included for comparison, including AutoGluon~\cite{tang2024autogluon}, AutoSklearn~\cite{feurer2022auto}, AIDE~\cite{Schmidt_Wu_Jiang}, Data Interpreter~\cite{hong2024data}, SELA~\cite{chi2024sela}.
Experiments are conducted based on the MetaGPT~\cite{hong2024metagpt} framework\footnote{\url{https://github.com/geekan/MetaGPT}}.
All datasets are split into training, validation, and test sets with a 6:2:2 ratio.
For the LLM-based Agents (I-MCTS, SELA, Data Interpreter, AIDE, etc), we choose Qwen2.5-72B-Instruct as the foundation LLM and employ a consistent initial task prompt across all methods~\footnote{Details can be refer to our code repository \url{https://anonymous.4open.science/r/I-MCTS}}. 
This prompt encompasses the dataset name, target column, and evaluation metric. 
We set the temperature parameter to 0.5. $\alpha_{\text{explore}}$ is set to 2. AIDE conducts 10 iterations per execution, while I-MCTS, SELA perform 10 rollouts. For I-MCTS and SELA, we employ Data Interpreter as the experimenter. The value of $\gamma$ in HRM is set to 0.2 across all datasets. This value is to control the decay rate of the LLM estimated score. Larger $\gamma$ makes the influence of the LLM estimated to persist longer, while smaller $\gamma$ allows the actual performance results to dominate the hybrid reward score more quickly. $\alpha_{explore}$ is set to 2. This aims to balance exploration and exploitation in the method’s search strategy. Following \citet{chi2024sela}, each method, except for AutoGluon, is run three times for each dataset. AutoGluon, being deterministic, is run only once with its default settings. AutoSklearn is also run with default settings, limited to 600 seconds per task.

\subsection{Evaluation Metrics}
\label{app:metric}
For the AMLB datasets, we use the default target column provided by OpenML. For Kaggle competition datasets, we rely on the target column specified in the competition description. Performance is measured using root mean squared error (RMSE) for regression tasks, F1 score for binary classification, and F1-weighted score for multi-class classification. To ensure comparability across datasets with varying metrics, we introduce a Normalized Score (NS), which maps RMSE into the range from 0 to 1.
\begin{align} 
    \text{NS}(s_{\text{raw}}) = \begin{cases} 
        \frac{1}{1 + \log{(1 + s_{\text{raw}})}} & \text{if the metric is RMSE.} \\
        s_{\text{raw}} & \text{otherwise.}
    \end{cases} 
\end{align}

\subsection{Main Results}

\begin{table}[t]
\centering
\resizebox{0.5\textwidth}{!}{
    \begin{tabular}{lccc}
        \toprule
        \textbf{Method} & \textbf{Top1 Rate} \%  & \textbf{Avg. NS} \% $\uparrow$ & \textbf{Avg. Best NS} \% $\uparrow$  \\
        \midrule
        AutoGluon   & 5.0 & 53.2 & 53.2 \\
        AutoSklearn  & 25.0 & 46.1 & 47.5  \\
        AIDE  & 5.0 & 47.1 & 51.8  \\
        Data Interpreter & 0.0 & 47.4 & 50.2 \\
        SELA & 20.0 & 55.4 & 55.9 \\
        \textbf{\methodname{}} & \textbf{45.0} & \textbf{58.6} & \textbf{59.8} \\
        \bottomrule
    \end{tabular}
    }
    \caption{Results of each AutoML framework on 20 tabular datasets. The ``Top1 Rate" column represents the rate of datasets where the method produces the best predictions across methods.}
    \label{table:main}
\end{table}

The experimental results demonstrate the superior performance of the proposed I-MCTS approach compared to other strong AutoML baselines. As shown in Table \ref{table:full-main-results}, I-MCTS achieves the highest overall normalized score (NS) of 58.6\% on average and 59.8\% for the best predictions across 20 diverse tabular datasets.
These results validate the effectiveness of I-MCTS's unique introspective node expansion process, which enables continuous solution refinement through meticulous analysis of parent and sibling nodes.
We include detailed results of each method in Appendix~\ref{app:detailed_result}.

\subsection{Ablation Study}

Performing ablation on all datasets is very time-consuming, hence, we employ a subset of datasets to do an ablation study. We follow \citet{chi2024sela} to choose the first two datasets alphabetically for each machine learning task type (Classification / Multi-class / Regression ). 
The results in Table \ref{table:ablation_study} illustrate the incremental benefits of incorporating both the Introspective Node Expansion (INE) and the hybrid reward mechanism (HRM).
Specifically, the results between I-MCTS (w/o HRM, rollout=5) and I-MCTS (rollout=5) shows that,
incorporating an LLM-estimated evaluation score facilitates better solutions to be explored earlier, especially under a lower resource scenario.


\begin{table}[t]
\centering
\resizebox{0.4\textwidth}{!}{
\begin{tabular}{lc}
    \toprule
    \textbf{Method} & \textbf{Avg. NS $\uparrow$}  \\
    \midrule
    Data Interpreter (r=10) & 56.4 \\
    SELA (Random Search, r=10) & 58.6  \\
    SELA (MCTS, r=10) & 60.9   \\
    \methodname{} (w/o INE, r=10) & 61.1  \\
    \methodname{} (w/o HRM, r=10) & 66.2  \\
    \methodname{} (r=10)  & 66.8  \\
    \midrule
    \methodname{} (w/o HRM, r=5) & 61.9  \\
    \methodname{} (r=5)  & 64.7  \\
    \bottomrule
\end{tabular}
}
\caption{Ablation Study for each search setting on the selected 6 datasets. "w/o INE" means "without "Introspective Node Expansion", while "w/o HRM" means ``without "hybrid reward mechanism". "r" means the rollout numbers.
}
\label{table:ablation_study}
\end{table}

\subsection{Cost-efficiency analysis}
\label{app:cost_efficiency analysis}

Our ablation studies across six datasets demonstrate that with rollouts=10, our method incurs about 30\% higher token consumption than SELA (averaging \$0.065 vs \$0.05 per dataset), which we consider comparatively equitable, given the empirical challenges in maintaining identical token consumption with baselines. 
More critically, we consider that experimental iteration counts constitute a more fundamental constraint than token budgets, as each agent trial necessitates substantial computational resource utilization (e.g., CPU/GPU allocation for model training). The relative significance of token costs will be anticipated to diminish with advancements in AI infrastructure and architectural optimizations.

\subsection{Case Study}

We conduct a comprehensive visualization analysis of the tree search process using the \textit{\textbf{kick}} dataset. 
As show in Appendix~\ref{app:case},
our empirical results reveal that I-MCTS effectively leverages the introspective information from previous nodes to generate task-specific and actionable insights, while SELA exhibit significant homogeneity and lack specificity.
Furthermore, our hybrid reward mechanism enhances the exploration efficiency, enabling more effective identification of high-quality nodes within the same computational budget.

\section{Conclusion}

In this paper, we introduced \textbf{I}ntrospective \textbf{M}onte \textbf{C}arlo \textbf{T}ree \textbf{S}earch (\textbf{\textit{I-MCTS}}), a novel approach to enhance AutoML Agents. 
The method addresses key limitations in existing Tree-search-based LLM agents with respect to thought diversity, quality, and the efficiency of the search process.
Our experimental results highlight the potential of integrating introspective capabilities into tree search algorithms for AutoML Agents.

\section*{Limitations}
While the proposed I-MCTS approach demonstrates significant improvements in AutoML agent performance, several limitations remain that warrant further investigation. First, the computational overhead of the introspective process, although beneficial for enhancing decision-making, introduces additional resource requirements. This limits the scalability of the method, particularly in scenarios where computational resources are constrained.
Future work should explore optimizations to reduce this overhead and investigate how the approach scales with increased computational power.
Also, our current implementation limits introspection to parent and sibling nodes to strike a balance between reflective depth and computational feasibility. Exploring deeper introspective hierarchies or incorporating a longer-term memory~\cite{xu2025mem} of the search trajectory are fascinating and important next steps.
Second, the current evaluation of I-MCTS is primarily focused on structured data and tabular ML tasks. Its effectiveness on more complex and heterogeneous data types, such as image and text data, remains unexplored. Extending the application of I-MCTS to these domains could provide valuable insights into its generalizability and robustness across diverse machine learning tasks, such as Vision, NLP and RL tasks.

\bibliography{custom}


\onecolumn
\appendix

\newpage
\newpage
\section{Algorithm}

\begin{algorithm}[H]
    \caption{Introspective Monte Carlo Tree Search (I-MCTS)}
    \begin{algorithmic}[1]
        \REQUIRE Problem description $p$, dataset $D$, metadata $d$, LLM $LLM$, rollout number $k$, blending constant $\gamma$.
        
        \STATE Initialize Tree with root node $x_{\text{root}}$
        \FOR{$i$ = 1 \textbf{to} $k$}
            \STATE node $x \leftarrow$ select(Tree) \COMMENT{Selection using UCT formula}
            \STATE $X_{\text{child}} \leftarrow$ expand(Tree, $x$, $x_{\text{parent}}$, $x_{\text{sibling}}$) \COMMENT{Introspective expansion}
            \STATE Choose the highest score node $x_{\text{sample}}$ from $X_{\text{child}}$
            \STATE Retrieve solution $\sigma_{\text{sol}}(x_{\text{sample}})$
            \STATE $\sigma_{\text{sol}}, s_{\text{actual}} \leftarrow \text{simulate}(\sigma_{\text{sol}}(x_{\text{sample}}), p, d, D, M)$
            \STATE $s_{\text{LLM}} \leftarrow LLM_{\text{value}}(\sigma_{\text{sol}}(x_{\text{sample}}))$
            \STATE $Q(x_{\text{sample}}) \leftarrow \alpha(x_{\text{sample}}) \cdot s_{\text{LLM}} + (1 - \alpha(x_{\text{sample}})) \cdot s_{\text{actual}}$
            \STATE Backpropagate(Tree, $Q(x_{\text{sample}})$)
        \ENDFOR
        \STATE $x_{\text{best}} \leftarrow \underset{x \in \text{Tree}}{\text{argmax}}(Q(x))$

        \ENSURE $\sigma_{\text{sol}}(x_{\text{best}})$
    \end{algorithmic}
    \label{alg:i-mcts}
\end{algorithm}

\section{Detailed Results}
\label{app:detailed_result}

Below show all detailed results on each dataset.

\begin{table}[h!]
\centering
\resizebox{\textwidth}{!}{
\begin{tabular}{l*{12}{c}} 
\toprule
 & \multicolumn{2}{c}{\textbf{AutoGluon}} & \multicolumn{2}{c}{\textbf{AutoSklearn}} & \multicolumn{2}{c}{\textbf{AIDE}} & \multicolumn{2}{c}{\textbf{DI}} & \multicolumn{2}{c}{\textbf{SELA}} & \multicolumn{2}{c}{\textbf{\methodname{}}} \\ 
\normalsize Dataset & {Avg.} & {Best} & {Avg.} & {Best} & {Avg.} & {Best} & {Avg.} & {Best} & {Avg.} & {Best} & {Avg.} & {Best} \\
\midrule
Click\_prediction\_small & 26.6 & 26.6 & 40.2 & 40.3 & 26.1 & 39.4 & 12.9 & 13.9 & 10.1 & 10.4 & 29.1 & 30.2 \\
GesturePhaseSegmentationProcessed & 69.3 & 69.3 & 67.2 & 68.4 & 56.3 & 68.1 & 60.1 & 64.4 & 63.0 & 63.4 & 67.2 & 68.5 \\
Moneyball & 24.3 & 24.3 & 13.1 & 13.8 & 23.8 & 24.6 & 9.5 & 24.5 & 41.2 & 41.6 & 41.6 & 41.7 \\
SAT11-HAND-runtime-regression & 12.6 & 12.6 & 10.3 & 10.3 & 12.0 & 12.1 & 11.4 & 11.9 & 23.3 & 23.7 & 21.7 & 24.6 \\
boston & 39.8 & 39.8 & 19.5 & 19.6 & 40.5 & 41.3 & 37.0 & 38.6 & 62.1 & 62.5 & 62.5 & 62.7 \\
colleges & 88.3 & 88.3 & 2.1 & 2.1 & 86.0 & 87.8 & 87.5 & 87.7 & 93.8 & 94.1 & 94.1 & 94.3 \\
concrete-strength & 28.3 & 28.3 & 17.4 & 17.9 & 28.3 & 28.3 & 28.8 & 29.6 & 46.8 & 47.3 & 47.3 & 48.2 \\
credit-g & 50.5 & 50.5 & 35.1 & 44.0 & 21.6 & 48.4 & 48.1 & 53.2 & 40.1 & 40.8 & 50.9 & 51.1 \\
diamonds & 13.8 & 13.8 & 8.7 & 8.7 & 13.7 & 13.7 & 13.5 & 13.6 & 26.4 & 26.7 & 26.8 & 26.8\\
house-prices & 9.0 & 9.0 & 2.0 & 2.0 & 8.9 & 8.9 & 8.5 & 9.0 & 17.9 & 18.2 & 15.8 & 18.4 \\
icr & 76.2 & 76.2 & 70.4 & 79.2 & 31.7 & 35.9 & 57.8 & 60.6 & 71.8 & 72.3 & 72.3 & 79.0 \\
jasmine & 84.3 & 84.3 & 84.4 & 84.7 & 83.6 & 84.6 & 77.8 & 83.5 & 83.0 & 83.5 & 84.2 & 84.3\\
kc1 & 38.3 & 38.3 & 43.5 & 45.0 & 40.8 & 42.6 & 38.1 & 41.2 & 38.1 & 38.6 & 44.8 & 45.6 \\
kick & 39.6 & 39.6 & 41.8 & 42.1 & 14.9 & 38.6 & 2.8 & 4.2 & 33.7 & 34.1 & 36.2 & 42.4 \\
mfeat-factors & 96.7 & 96.7 & 97.1 & 97.5 & 94.4 & 94.5 & 93.0 & 96.0 & 95.0 & 95.2 & 94.4 & 95.4\\
segment & 93.5 & 93.5 & 92.7 & 93.1 & 91.7 & 92.2 & 91.7 & 92.6 & 93.2 & 93.5 & 91.1 & 93.7\\
smoker-status & 78.0 & 78.0 & 78.6 & 78.9 & 74.8 & 76.3 & 77.3 & 81.5 & 85.8 & 86.4 & 79.9 & 91.5\\
software-defects & 51.5 & 51.5 & 61.1 & 61.7 & 49.7 & 49.8 & 54.5 & 57.3 & 53.2 & 53.7 & 60.5 & 60.7\\
titanic & 78.9 & 78.9 & 76.2 & 78.9 & 81.2 & 83.7 & 76.0 & 78.5 & 68.0 & 68.6 & 76.1 & 78.5\\
wine-quality-white & 65.4 & 65.4 & 60.7 & 61.4 & 62.9 & 65.1 & 61.2 & 61.6 & 62.2 & 62.6 & 63.3 & 65.5 \\
\midrule
\normalsize Overall NS \% $\uparrow$ & 53.2 & 53.2 & 46.1 & 47.5  & 45.5 & 51.8 & 47.4 & 50.2 & 55.4 & 55.9 & \textbf{58.6} & \textbf{59.8} \\
\bottomrule
\end{tabular}
}
\caption{Methods' NS \% for each tabular dataset}
\label{table:full-main-results}
\end{table}

\newpage

\section{Datasets}
\label{app:dataests}

Table \ref{table:datasets} outlines the detailed information of the datasets used for evaluation. For ablation study, we use boston, colleges, credit-g, Click\_prediction\_small, GesturePhaseSegmentationProcessed, and mfeat-factors. 

\begin{table*}[h]
\centering
\small
\resizebox{.99\textwidth}{!}{
\begin{tabular}{>{\scriptsize}p{0.25\textwidth}>{\scriptsize}p{0.1\textwidth}<{\centering}>{\scriptsize}c>{\scriptsize}c>{\scriptsize}c>{\scriptsize}c>{\scriptsize}c}

\toprule
\footnotesize \textbf{Dataset name}    & \footnotesize \textbf{\# Features} & \footnotesize \textbf{\# Rows}  & \footnotesize \textbf{\# Classes} & \footnotesize \textbf{Task Type}  &  \footnotesize \textbf{Metric}  & \footnotesize \textbf{Source}   \\
\midrule
boston          & 14          & 506      & N/A        & Regression & RMSE  & OpenML \scriptsize (Dataset ID: 531)     \\
colleges                                  & 48          & 7063     & N/A        & Regression & RMSE   & OpenML \scriptsize (Dataset ID: 42727)     \\
concrete-strength        & 9           & 4866     & N/A        & Regression & RMSE  & Kaggle \scriptsize (playground-series-s3e9)      \\
diamonds                                  & 10          & 53940    & N/A        & Regression & RMSE  & OpenML \scriptsize (Dataset ID: 42225)     \\
house-prices                               & 81          & 1460     & N/A        & Regression & RMSE & Kaggle  \scriptsize (house-prices-advanced-regression-techniques)     \\
Moneyball                                 & 15          & 1232     & N/A        & Regression & RMSE  & OpenML \scriptsize (Dataset ID: 41021)      \\
SAT11-HAND-runtime-regression             & 118         & 4440     & N/A        & Regression & RMSE  & OpenML \scriptsize (Dataset ID: 41980)    \\
credit-g                                  & 21          & 1000     & 2          & Classification & F1 & OpenML \scriptsize (Dataset ID: 31)      \\
Click\_prediction\_small                    & 12          & 39948    & 2          & Classification & F1  & OpenML \scriptsize (Dataset ID: 42733)    \\

icr   & 58          & 617      & 2          & Classification & F1 & Kaggle  \scriptsize (icr-identify-age-related-conditions)     \\
jasmine                                   & 145         & 2984     & 2          & Classification & F1 & OpenML \scriptsize (Dataset ID: 41143)       \\
kc1                                       & 21          & 2109     & 2          & Classification & F1  & OpenML \scriptsize (Dataset ID: 1067)     \\
kick                                      & 33          & 72983    & 2          & Classification & F1  & OpenML \scriptsize (Dataset ID: 41162)     \\
smoker-status                             & 23          & 143330   & 2          & Classification & F1  & Kaggle \scriptsize (playground-series-s3e24)     \\
software-defects                          & 22          & 91586    & 2          & Classification & F1  & Kaggle \scriptsize (playground-series-s3e23)     \\
titanic                                   & 12          & 891      & 2          & Classification & F1  & Kaggle \scriptsize (titanic)    \\
GesturePhaseSegmentationProcessed      & 33          & 9873     & 5          & Multiclass & F1-weighted & OpenML \scriptsize (Dataset ID: 4538) \\

mfeat-factors                             & 217         & 2000     & 10         & Multiclass & F1-weighted  & OpenML \scriptsize (Dataset ID: 12)\\

segment                                   & 20          & 2310     & 7          & Multiclass & F1-weighted & OpenML \scriptsize (Dataset ID: 40984) \\
wine-quality-white                      & 12          & 4898     & 7          & Multiclass & F1-weighted & OpenML \scriptsize (Dataset ID: 40498) \\
\bottomrule
\hline
\end{tabular}
}
\caption{Summary of the machine learning datasets used in the experiments. OpenML datasets can be accessed using their respective dataset IDs. The Kaggle datasets are available at https://www.kaggle.com/competitions/\{source\}.}
\label{table:datasets}
\end{table*}

\section{Prompts}
\label{app:prompt}

\subsection{Introspection Node Expansion Prompt}
\begin{lstlisting}[style=pythonstyle]
INTROSPECTION_PROMPT = """
You are an AI assistant tasked with analyzing a machine learning solution and proposing new insights to IMPROVE its performance. Given the current plan, solution code, Evaluation Metrics and performance on development score, suggest innovative approaches to ENHANCE and IMPROVE the performance.

# Current Plan:
{current_plan}

# Current Solution Code:
{solution_code}

# Current Evaluation Metrics:
{current_metrics}

# Current Performance on Development Set: 
{dev_score}

There may be some drawbacks in the Current Plan or Solution that leads to a lower development score.
Based on this information, please propose ONE new insight w.r.t. the following specific stage of the machine learning pipeline to IMPROVE the overall performance. Your insight should be specific, actionable, and have the potential to IMPROVE the model's performance.
{task_type}

NOTE that the following insights have been proposed before. Kindly AVOID suggesting an insight that is similar to them.
{cur_expansion_list}

Please strictly format your response as a JSON with the following structure:
```json
{{
    "task_type": "Data Preprocessing",
    "critic_feedback": "feedback and explanation on potential drawbacks in the Current Plan or Solution that leads to a lower development score"
    "insight": "ONE insight that specific, actionable, and have the potential to IMPROVE the model's performance."
}}
```
"""
\end{lstlisting}

\subsection{Node Evaluation Prompt}
Each Node will be evaluated by the LLM based on the following prompt.
\label{app:llm_evaluation_schema}

\begin{lstlisting}[style=pythonstyle]
EVALUATOR_SYSTEM_MSG = "As a Kaggle Grandmaster competing in a challenge, Your role is to evaluate the candidate solution for the given data science problem."

EVALUATION_CRETERIA = """
### Evaluation Criteria for AI Engineer's Execution Plan in Kaggle Data Science Competition  
Below is a scoring framework to evaluate the quality and feasibility of an execution plan. The total score is **100 points**, divided into **7 criteria**, each with defined scoring ranges and descriptions.  
---
#### **1. Problem Understanding (15 points)**  
- **15**: Clearly defines the competition goal, success metrics (e.g., AUC, RMSE), and constraints (e.g., runtime limits). Includes domain-specific nuances (e.g., business impact for sales prediction).  
- **10 to 14**: Adequate understanding but misses minor details (e.g., unclear evaluation metric implications).  
- **5 to 9**: Superficial analysis; overlooks key competition rules or data constraints.  
- **0 to 4**: Incorrect problem framing or missing critical objectives.  
---
#### **2. Data Preprocessing & Feature Engineering (20 points)**  
- **20**: Comprehensive plan addressing missing values, outliers, categorical encoding, and normalization. Proposes novel features (e.g., interaction terms, domain-specific transformations) and validates their impact.  
- **15 to 19**: Solid strategy but lacks innovation (e.g., standard scaling but no feature interactions).  
- **10 to 14**: Basic techniques (e.g., mean imputation, one-hot encoding) with gaps (e.g., no outlier handling).  
- **5 to 9**: Incomplete or naive methods (e.g., dropping all missing values without analysis).  
- **0 to 4**: No preprocessing/feature engineering or harmful approaches.  
---
#### **3. Model Selection & Validation Strategy (20 points)**  
- **20**: Justifies model choices (e.g., LightGBM for tabular data, NN for high dimensionality) and advanced techniques (e.g., stacking, automated hyperparameter tuning). Designs robust cross-validation aligned with competition rules (e.g., time-series splits for temporal data).  
- **15 to 19**: Reasonable models (e.g., XGBoost) and CV but lacks optimization (e.g., no Bayesian hyperparameter search).  
- **10 to 14**: Basic models (e.g., random forest) with weak validation (e.g., simple holdout).  
- **5 to 9**: Inappropriate models (e.g., CNNs for small tabular data) or validation leakage.  
- **0 to 4**: No clear model selection or validation plan.  
---
#### **4. Training & Optimization (15 points)**  
- **15**: Efficient resource use (e.g., GPU acceleration, parallel trials), advanced optimization (e.g., class imbalance handling, quantile loss for skewed targets), and time management.  
- **10 to 14**: Logical training workflow but suboptimal resource allocation (e.g., no early stopping).  
- **5 to 9**: Basic training loop with critical inefficiencies (e.g., no hyperparameter tuning).  
- **0 to 4**: Unworkable training strategy (e.g., no batch processing for large data).  
---
#### **5. Post-Processing & Interpretation (10 points)**  
- **10**: Ensemble methods (e.g., weighted blending), thorough error analysis, and model interpretation (e.g., SHAP values, feature importance).  
- **7 to 9**: Simple ensembling (e.g., averaging) and basic interpretation (e.g., feature importance plots).  
- **4 to 6**: Minimal post-processing (e.g., no calibration) or superficial analysis.  
- **0 to 3**: No post-processing or interpretation.  
---
#### **6. Documentation & Reproducibility (10 points)**  
- **10**: Clean, modular code structure with detailed documentation (e.g., Docker setup, dependency lists). Includes version control and experiment tracking (e.g., MLflow).  
- **7 to 9**: Readable code but limited documentation (e.g., no README).  
- **4 to 6**: Disorganized code with critical reproducibility gaps.  
- **0 to 3**: Undocumented or non-reproducible code.  
---
#### **7. Innovation & Practicality (10 points)**  
- **10**: Novel yet practical ideas (e.g., synthetic data for imbalance, custom loss functions). Balances creativity with competition constraints.  
- **7 to 9**: Minor innovations (e.g., new feature engineering) but overly complex in parts.  
- **4 to 6**: Generic approach with no novel elements.  
- **0 to 3**: Impractical or gimmicky methods (e.g., unnecessary deep learning).  
---
### **Scoring Scale**  
| Total Score | Grade          | Description  |  
| **90 to 100**  | Exceptional    | Well-structured, innovative, and executable plan. High chance of top ranks. |  
| **70 to 89**   | Strong         | Minor gaps but logically sound. Competitive but needs refinement.           |  
| **50 to 69**   | Average        | Basic approach with significant oversights (e.g., weak validation).         |  
| **<50**     | Insufficient   | Flawed or incomplete plan; unlikely to succeed.                             |  
---
### **Evaluation Guidance**  
- Prioritize **justification of choices** (e.g., why CatBoost over XGBoost?).  
- Reward **pragmatism** (e.g., focusing on feature engineering over overly complex models).  
- Penalize **ignoring competition constraints** (e.g., submission runtime limits).  
- Highlight **risks** (e.g., overfitting due to improper CV) and mitigation strategies.  
This framework ensures a balance between technical rigor, innovation, and practicality in tabular data competitions.
"""

EVALUATE_NODE_BY_LLM = """
# Evaluation Criteria
{evaluation_Criteria}
# Instruction:
Based on the above evaluation Criteria, Please give feedback and rating for the following solution. 
# Candidate Case
## User Requirement
{user_requirement}
## Candidate Solution
{candidate_plan}
# Format:
Please ensure your output strictly adheres to the following structure, ensure the "Total_Score" is a INT number":
```json
{{
    "evaluation_feedback": "Your evaluation feedback for the candidate solution based on the evaluation criteria. Please provide a detailed analysis of each criterion and explain why you gave the specific rating",
    "total_score": x
}}
\end{lstlisting}

\section{Case Study}
\label{app:case}
We detail the case study below. The refined insight generated by I-MCTS suggests that "Developing a feature that captures the temporal difference between the vehicle's manufacturing year and purchase date (PurchDate-VehYear) would enable more precise representation of the vehicle's age at the time of purchase.", which is more task-specific.
In contrast, the insights generated by SELA exhibit significant homogeneity and lack specificity. 
\subsection{Overview of I-MCTS's search process}

\begin{lstlisting}[style=txtfile]
Number of simulations: 10
Node id: 0
Plans: 
1. Perform exploratory data analysis (EDA) on the training dataset to understand the distribution of features and identify any missing values or outliers.
2. Preprocess the data by handling missing values, encoding categorical variables, and normalizing numerical features. Ensure that the preprocessing steps are applied consistently to the train, dev, and test sets.
3. Create new features that might improve the model's performance, such as interaction terms or polynomial features, without using the target column.
4. Train a machine learning model on the preprocessed training data. Use the dev set to tune hyperparameters and validate the model's performance.
5. Evaluate the model's performance on the training, dev, and test sets using the F1 score. Save the predictions for the dev and test sets in the specified format.
Simulated: True
Score: avg score: 0.17933643065796115, simulated score: {'test_score': 0.08495018353434715, 'dev_score': 0.05279187817258883, 'score': 0.05279187817258883}, Visits: 10

	Node id: 0-0
	Plans: 
	3. Create interaction features between numerical and categorical variables. For example, you can create new features by multiplying numerical features with the encoded values of categorical features. This can help the model capture interactions that might be important for predicting the target variable. Additionally, consider creating polynomial features for numerical columns to capture non-linear relationships. For instance, you can add squared and cubed terms of numerical features to the dataset.
	Simulated: True
	Score: avg score: 0.19208087615838249, simulated score: {'test_score': 0, 'dev_score': 0, 'score': 0}, Visits: 2

		Node id: 0-0-0
		Plans: 
		4. Implement an ensemble of models using techniques such as stacking or boosting. For example, use a combination of Gradient Boosting Machines (GBM), Random Forests, and XGBoost. This can help capture different aspects of the data and improve the robustness of the model. Additionally, use cross-validation to tune hyperparameters for each model in the ensemble, ensuring that the model generalizes well to unseen data.
		Simulated: False
		Score: avg score: 0, simulated score: {}, Visits: 0

		Node id: 0-0-1
		Plans: 
		4. Introduce a neural network model with multiple hidden layers to capture complex interactions and non-linear relationships in the data. Use techniques like dropout and batch normalization to prevent overfitting. Additionally, experiment with different activation functions (e.g., ReLU, LeakyReLU) and loss functions (e.g., binary cross-entropy) to find the best configuration. Use early stopping to prevent overfitting and ensure the model generalizes well to the dev set.
		Simulated: False
		Score: avg score: 0, simulated score: {}, Visits: 0

		Node id: 0-0-2
		Plans: 
		4. Implement a Bayesian Optimization approach for hyperparameter tuning. Bayesian Optimization is a sequential design strategy for global optimization of black-box functions, which can be particularly effective in finding the optimal hyperparameters for machine learning models. This method can help in exploring the hyperparameter space more efficiently and effectively, leading to better model performance. Use a library like Optuna or Hyperopt to automate the process and integrate it with cross-validation to ensure that the model generalizes well to unseen data.
		Simulated: False
		Score: avg score: 0, simulated score: {}, Visits: 0

		Node id: 0-0-3
		Plans: 
		4. Implement a hybrid model that combines a traditional machine learning model (e.g., Random Forest or XGBoost) with a neural network. The traditional model can capture the linear and tree-based relationships, while the neural network can capture the complex and non-linear interactions. Use the predictions from the traditional model as additional features (meta-features) for the neural network. This hybrid approach can help the model capture a broader range of patterns in the data, potentially leading to improved performance on the development set.
		Simulated: False
		Score: avg score: 0, simulated score: {}, Visits: 0

		Node id: 0-0-4
		Plans: 
		4. Implement a feature selection process using Recursive Feature Elimination (RFE) or permutation importance to identify and retain the most relevant features for the model. This can help reduce overfitting by removing noise and irrelevant features, leading to a more robust model. Additionally, use a cross-validated feature selection approach to ensure that the selected features generalize well to unseen data. This can improve the model's performance on the development set by focusing on the most predictive features.
		Simulated: True
		Score: avg score: 0.38416175231676497, simulated score: {'train_score': 0.9999073645206114, 'dev_score': 0.38416175231676497, 'test_score': 0.36547776309995594, 'score': 0.38416175231676497}, Visits: 1

	Node id: 0-1
	Plans: 
	3. Introduce domain-specific feature engineering by creating new features that are relevant to the automotive industry. For example, you can create a feature that calculates the age of the vehicle at the time of purchase (PurchDate - VehYear), which might be a more meaningful representation of vehicle age than the current 'VehicleAge' feature. Additionally, consider creating a feature that represents the difference between the current and acquisition MMR prices (e.g., MMRCurrentAuctionAveragePrice - MMRAcquisitionAuctionAveragePrice), which could capture the depreciation or appreciation of the vehicle's value. These domain-specific features can provide the model with more relevant information and potentially improve its performance.
	Simulated: False
	Score: avg score: 0.0, simulated score: {'test_score': 0, 'dev_score': 0, 'score': 0}, Visits: 1

	Node id: 0-2
	Plans: 
	3. Introduce time-based features to capture temporal trends and seasonality. For example, extract the month and day of the week from the 'PurchDate' column. This can help the model understand if certain times of the year or days of the week have a significant impact on the likelihood of a bad buy. Additionally, consider creating a feature that represents the time difference between the vehicle's year and the purchase date (PurchDate - VehYear) to capture the age of the vehicle at the time of purchase more accurately. These time-based features can provide the model with additional context and potentially improve its performance.
	Simulated: True
	Score: avg score: 0.41769512423377475, simulated score: {'test_score': 0.37755545889517184, 'dev_score': 0.39700996677740863, 'score': 0.39700996677740863}, Visits: 2

		Node id: 0-2-0
		Plans: 
		4. Implement an ensemble of models, such as a Random Forest and a Gradient Boosting Machine (GBM), and use stacking to combine their predictions. This approach can help capture different aspects of the data and improve the robustness of the model. Additionally, use a more systematic hyperparameter tuning method, such as Bayesian optimization, to find the best settings for each model in the ensemble.
		Simulated: False
		Score: avg score: 0, simulated score: {}, Visits: 0

		Node id: 0-2-1
		Plans: 
		4. Implement a more sophisticated model training strategy by using a deep learning model, such as a neural network with multiple layers, to capture complex feature interactions. Additionally, incorporate techniques to handle class imbalance, such as oversampling the minority class or using class weights during training. Use K-fold cross-validation to ensure that the model generalizes well to unseen data and to avoid overfitting. Finally, experiment with different activation functions and regularization techniques to improve model performance.
		Simulated: False
		Score: avg score: 0, simulated score: {}, Visits: 0

		Node id: 0-2-2
		Plans: 
		4. Implement a more advanced model training strategy by using a neural network with attention mechanisms. Attention mechanisms can help the model focus on the most relevant features and capture complex interactions between features. Additionally, use a learning rate scheduler to dynamically adjust the learning rate during training, which can help the model converge more effectively and avoid getting stuck in local minima. Finally, incorporate early stopping to prevent overfitting by monitoring the performance on the development set and stopping training when performance starts to degrade.
		Simulated: False
		Score: avg score: 0, simulated score: {}, Visits: 0

		Node id: 0-2-3
		Plans: 
		4. Implement a more advanced model training strategy by using a neural network with a combination of convolutional and recurrent layers. Convolutional layers can help capture local patterns and feature interactions, while recurrent layers can handle sequential data and temporal dependencies. Use a learning rate scheduler to dynamically adjust the learning rate during training, which can help the model converge more effectively and avoid getting stuck in local minima. Additionally, incorporate dropout layers to prevent overfitting and improve the model's generalization ability. Finally, use K-fold cross-validation to ensure that the model generalizes well to unseen data and to avoid overfitting.
		Simulated: False
		Score: avg score: 0, simulated score: {}, Visits: 0

		Node id: 0-2-4
		Plans: 
		4. Implement a more advanced model training strategy by using a neural network with a combination of dense and embedding layers. Use embeddings for categorical variables to capture their latent representations and dense layers to model complex interactions between features. Incorporate techniques to handle class imbalance, such as focal loss, which assigns higher weights to misclassified samples, particularly from the minority class. Use a learning rate scheduler to dynamically adjust the learning rate during training, and incorporate early stopping to prevent overfitting. Finally, use K-fold cross-validation to ensure that the model generalizes well to unseen data and to avoid overfitting.
		Simulated: True
		Score: avg score: 0.43838028169014087, simulated score: {'test_score': 0.4241112828438949, 'dev_score': 0.43838028169014087, 'score': 0.43838028169014087}, Visits: 1

	Node id: 0-3
	Plans: 
	3. Introduce feature engineering based on domain knowledge and feature interactions. For example, create a feature that represents the ratio of the vehicle's odometer reading to its age (VehOdo / VehicleAge), which could provide insights into the vehicle's usage intensity. Additionally, consider creating interaction features between categorical and numerical variables, such as the interaction between the vehicle's make and its MMR prices (e.g., Make * MMRCurrentAuctionAveragePrice). These features can help the model capture more nuanced relationships and potentially improve its performance.
	Simulated: True
	Score: avg score: 0.036288232244686365, simulated score: {'test_score': 0.07784120394395433, 'dev_score': 0.036288232244686365, 'score': 0.036288232244686365}, Visits: 2

		Node id: 0-3-0
		Plans: 
		4. Implement a more robust hyperparameter tuning strategy using techniques like Grid Search or Randomized Search. This will help in finding the optimal set of hyperparameters that can improve the model's performance on the development set. Additionally, consider using cross-validation during the hyperparameter tuning process to ensure that the model generalizes well to unseen data. For example, you can use `GridSearchCV` or `RandomizedSearchCV` from `scikit-learn` to systematically explore different hyperparameter combinations and select the best model based on the F1 score on the cross-validation sets.
		Simulated: False
		Score: avg score: 0, simulated score: {}, Visits: 0

		Node id: 0-3-1
		Plans: 
		4. Implement an ensemble of models to improve the robustness and generalization of the predictions. Specifically, use a stacking ensemble where multiple base models (e.g., Random Forest, Gradient Boosting, and Logistic Regression) are trained and their predictions are combined using a meta-model (e.g., another Logistic Regression or a simple averaging). This approach can help capture different aspects of the data and reduce the risk of overfitting. Additionally, use Bayesian optimization for hyperparameter tuning to systematically explore the hyperparameter space and find the optimal settings for each base model and the meta-model.
		Simulated: False
		Score: avg score: 0, simulated score: {}, Visits: 0

		Node id: 0-3-2
		Plans: 
		4. Implement class weighting or oversampling techniques to address class imbalance in the training data. Since the target variable 'IsBadBuy' is likely imbalanced, using class weights in the loss function or oversampling the minority class can help the model learn from the underrepresented class more effectively. For example, in scikit-learn, you can set the `class_weight` parameter in the model to 'balanced' or use the `imbalanced-learn` library to perform oversampling techniques like SMOTE (Synthetic Minority Over-sampling Technique). This can improve the model's ability to predict the minority class, which is crucial for the F1 score.
		Simulated: False
		Score: avg score: 0, simulated score: {}, Visits: 0

		Node id: 0-3-3
		Plans: 
		4. Implement advanced regularization techniques such as L1 (Lasso) and L2 (Ridge) regularization to prevent overfitting and improve the model's generalization. Specifically, use Elastic Net regularization, which combines both L1 and L2 penalties, to balance the trade-off between feature selection and coefficient shrinkage. This can help in reducing the complexity of the model and improving its performance on the development set. Additionally, consider using dropout regularization if you are using neural networks, which randomly drops out (sets to zero) a number of output features of the layer during training, thus preventing the model from becoming too reliant on any single feature.
		Simulated: False
		Score: avg score: 0, simulated score: {}, Visits: 0

		Node id: 0-3-4
		Plans: 
		4. Implement a more sophisticated model architecture that can capture complex interactions and non-linear relationships, such as a Gradient Boosting Machine (GBM) or a Deep Neural Network (DNN). For GBM, use a library like XGBoost or LightGBM, which are known for their efficiency and performance in handling large datasets and capturing complex patterns. For DNN, use a multi-layer perceptron with appropriate activation functions (e.g., ReLU) and regularization techniques (e.g., dropout) to prevent overfitting. Additionally, use learning rate schedules and early stopping to optimize the training process and prevent overfitting. This can help the model better capture the nuances in the data and improve its performance on the development set.
		Simulated: False
		Score: avg score: 0.036288232244686365, simulated score: {'test_score': 0.07784120394395433, 'dev_score': 0.036288232244686365, 'score': 0.036288232244686365}, Visits: 1

	Node id: 0-4
	Plans: 
	3. Introduce feature engineering using dimensionality reduction techniques such as Principal Component Analysis (PCA) or t-Distributed Stochastic Neighbor Embedding (t-SNE) to create new features that capture the underlying structure of the data. For example, apply PCA to the numerical features to reduce dimensionality and create a set of principal components that can be used as new features. This can help in capturing the most important information in the data while reducing noise and multicollinearity. Additionally, consider using autoencoders to learn a compressed representation of the data, which can be used as input features for the model. These techniques can help in uncovering hidden patterns and improving the model's performance.
	Simulated: True
	Score: avg score: 0.22422198156666764, simulated score: {'test_score': 0.3816859221223791, 'dev_score': 0.38071487946799665, 'score': 0.38071487946799665}, Visits: 2

		Node id: 0-4-0
		Plans: 
		4. Implement an ensemble of models to improve the robustness and performance of the predictions. Specifically, use a combination of different algorithms such as Random Forest, Gradient Boosting Machines (GBM), and XGBoost. Train each model on the preprocessed data and combine their predictions using techniques like stacking or voting. This approach can help in capturing different patterns in the data and reducing overfitting, leading to better generalization and higher F1 scores on the development set.
		Simulated: False
		Score: avg score: 0, simulated score: {}, Visits: 0

		Node id: 0-4-1
		Plans: 
		4. Introduce a neural network model with dropout regularization to improve the model's ability to generalize. Specifically, use a multi-layer perceptron (MLP) with dropout layers to prevent overfitting. Dropout randomly sets a fraction of input units to 0 at each update during training, which helps in reducing overfitting and improving the model's performance on unseen data. Additionally, use early stopping to monitor the performance on the dev set and stop training when the performance starts to degrade, ensuring that the model does not overfit to the training data. This approach can help in capturing complex interactions in the data and improving the F1 score on the development set.
		Simulated: False
		Score: avg score: 0, simulated score: {}, Visits: 0

		Node id: 0-4-2
		Plans: 
		4. Introduce a deep learning model with attention mechanisms to improve the model's ability to capture complex and non-linear relationships in the data. Specifically, use a Transformer-based model or a model with self-attention layers. These models can dynamically weigh the importance of different features and interactions, which can be particularly useful for datasets with a large number of features and complex relationships. This approach can help in improving the model's performance on the development set by better capturing the underlying patterns in the data.
		Simulated: False
		Score: avg score: 0, simulated score: {}, Visits: 0

		Node id: 0-4-3
		Plans: 
		4. Implement a grid search with cross-validation to systematically tune hyperparameters for the model. Specifically, use GridSearchCV from scikit-learn to perform an exhaustive search over a specified parameter grid. This will help in finding the optimal hyperparameters for the model, which can significantly improve its performance. Additionally, consider using K-Fold cross-validation to ensure that the model is robust and generalizes well to unseen data. This approach can help in reducing overfitting and improving the F1 score on the development set.
		Simulated: False
		Score: avg score: 0, simulated score: {}, Visits: 0

		Node id: 0-4-4
		Plans: 
		4. Introduce a hybrid model that combines a traditional machine learning model (e.g., Random Forest) with a deep learning model (e.g., a Convolutional Neural Network, CNN). The CNN can be used to extract high-level features from the numerical and categorical data, which can then be concatenated with the features from the Random Forest. This hybrid approach can help in capturing both linear and non-linear relationships in the data, leading to improved performance. Specifically, use the CNN to process the numerical features and the Random Forest to handle the categorical features, and then combine their outputs using a fully connected layer. This approach can help in improving the F1 score on the development set by leveraging the strengths of both models.
		Simulated: True
		Score: avg score: 0.06772908366533864, simulated score: {'train_score': -1, 'dev_score': 0.06772908366533864, 'test_score': 0.061683220073183484, 'score': 0.06772908366533864}, Visits: 1

Generated 30 unique codes.
Best node: 0-2-4, score: {'test_score': 0.4241112828438949, 'dev_score': 0.43838028169014087, 'score': 0.43838028169014087}
Dev best node: 0-2-4, score: {'test_score': 0.4241112828438949, 'dev_score': 0.43838028169014087, 'score': 0.43838028169014087}
Grader score: 0.4241112828438949
\end{lstlisting}

\subsection{Overview of baseline SELA's search process}

\begin{lstlisting}[style=txtfile]
Number of simulations: 10
[Node 0]
Plans: 
1. Perform exploratory data analysis (EDA) on the training dataset to understand the data distribution, missing values, and potential outliers.
2. Preprocess the data by handling missing values, encoding categorical variables, and normalizing numerical features.
3. Create new features that might improve the model's performance, such as interaction terms or domain-specific features.
4. Train a machine learning model using the preprocessed and feature-engineered training data.
5. Evaluate the model's performance on the development set using the F1 score.
6. Make predictions on the test set and save the results to `test_predictions.csv`.
7. Make predictions on the development set and save the results to `dev_predictions.csv`.
8. Save the trained model to `model.pkl`.
9. Print the performance metrics (F1 scores) for the training, development, and test sets.
Simulated: True
Score: avg score: 0.13309993361366776, simulated score: {'test_score': 0.1530317613089509, 'dev_score': 0.14690841469084145, 'score': 0.14690841469084145}, Visits: 10

	[Node 0-0]
	Plans: 
	3. Before creating new features such as interaction terms or domain-specific features, analyze the distribution of 'IsBadBuy' to understand the class imbalance. This analysis should be performed on the training set after splitting the data to ensure that the insights gained are representative of the data the model will be trained on. Use these insights to guide the creation of new features that might improve the model's performance.
	Simulated: True
	Score: avg score: 0.14587155963302753, simulated score: {'test_score': 0.15175644028103044, 'dev_score': 0.14587155963302753, 'score': 0.14587155963302753}, Visits: 2

		[Node 0-0-0]
		Plans: 
		4. Train a machine learning model using the preprocessed and feature-engineered training data. Before training, analyze the distribution of 'IsBadBuy' to understand the class imbalance. If necessary, apply appropriate techniques to handle the imbalance during the data split and model training.
		Simulated: False
		Score: avg score: 0.14587155963302753, simulated score: {'test_score': 0.15175644028103044, 'dev_score': 0.14587155963302753, 'score': 0.14587155963302753}, Visits: 1

		[Node 0-0-1]
		Plans: 
		4. Train a machine learning model using the preprocessed and feature-engineered training data. Before training, explore the correlation between 'VehicleAge' and 'IsBadBuy' to determine if older vehicles are more likely to be classified as 'IsBadBuy'. Ensure that this exploration is conducted on the training set if the data is split into training and testing sets.
		Simulated: False
		Score: avg score: 0, simulated score: {}, Visits: 0

		[Node 0-0-2]
		Plans: 
		4. Train a machine learning model using the preprocessed and feature-engineered training data. Before training, investigate the relationship between 'MMRAcquisitionAuctionAveragePrice' and 'MMRAcquisitionAuctionCleanPrice' with 'IsBadBuy' to understand the pricing impact. Ensure that this analysis is conducted on the training data split to avoid data leakage.
		Simulated: False
		Score: avg score: 0, simulated score: {}, Visits: 0

		[Node 0-0-3]
		Plans: 
		4. Train a machine learning model using the preprocessed and feature-engineered training data. Before training, examine the frequency of 'Make' and 'Model' in the training data to identify which brands or models are more prone to being kicks. Ensure that this analysis is performed on the training set if the data is split.
		Simulated: False
		Score: avg score: 0, simulated score: {}, Visits: 0

		[Node 0-0-4]
		Plans: 
		4. Train a machine learning model using the preprocessed and feature-engineered training data. Before training, visualize the distribution of 'VehOdo' to understand if high mileage vehicles are more likely to be problematic. Ensure that the data is split appropriately for training and validation, and integrate the insights from the 'VehOdo' distribution into the model training process.
		Simulated: False
		Score: avg score: 0, simulated score: {}, Visits: 0

	[Node 0-1]
	Plans: 
	3. Create new features that might improve the model's performance, such as interaction terms or domain-specific features. Specifically, explore the correlation between 'VehicleAge' and 'IsBadBuy' to see if older vehicles are more likely to be classified as 'kicks'. Consider adding interaction terms or derived features based on this relationship, and ensure that any new features are integrated into the model training process, including any data splits you perform.
	Simulated: True
	Score: avg score: 0.14587155963302753, simulated score: {'test_score': 0.15175644028103044, 'dev_score': 0.14587155963302753, 'score': 0.14587155963302753}, Visits: 2

		[Node 0-1-0]
		Plans: 
		4. Train a machine learning model using the preprocessed and feature-engineered training data. Before training, analyze the distribution of 'IsBadBuy' to understand the class imbalance, and ensure that any data splitting maintains the distribution of the target variable to avoid bias in the training and validation sets.
		Simulated: False
		Score: avg score: 0, simulated score: {}, Visits: 0

		[Node 0-1-1]
		Plans: 
		4. Train a machine learning model using the preprocessed and feature-engineered training data. Before training, explore the correlation between 'VehicleAge' and 'IsBadBuy' to determine if older vehicles are more likely to be classified as 'kicks'. Ensure that this exploration is conducted on the training data split to avoid data leakage.
		Simulated: False
		Score: avg score: 0.14587155963302753, simulated score: {'test_score': 0.15175644028103044, 'dev_score': 0.14587155963302753, 'score': 0.14587155963302753}, Visits: 1

		[Node 0-1-2]
		Plans: 
		4. Train a machine learning model using the preprocessed and feature-engineered training data. Before training, investigate the relationship between 'MMRAcquisitionAuctionAveragePrice' and 'MMRAcquisitionAuctionCleanPrice' with 'IsBadBuy' to understand the pricing impact. Ensure that this analysis is conducted on the training data split to avoid data leakage.
		Simulated: False
		Score: avg score: 0, simulated score: {}, Visits: 0

		[Node 0-1-3]
		Plans: 
		4. Train a machine learning model using the preprocessed and feature-engineered training data. Before training, examine the frequency of 'Make' and 'Model' in the dataset to identify which brands or models are more prone to being kicks. Ensure that this analysis is performed on the training set if the data is split into training and testing sets.
		Simulated: False
		Score: avg score: 0, simulated score: {}, Visits: 0

		[Node 0-1-4]
		Plans: 
		4. Before training the machine learning model using the preprocessed and feature-engineered training data, visualize the distribution of 'VehOdo' to understand if high mileage vehicles are more likely to be problematic. Ensure that this visualization is performed on the training set after splitting the data into training and validation sets.
		Simulated: False
		Score: avg score: 0, simulated score: {}, Visits: 0

	[Node 0-2]
	Plans: 
	3. Create new features that might improve the model's performance, such as interaction terms or domain-specific features. Specifically, investigate the relationship between 'MMRAcquisitionAuctionAveragePrice' and 'MMRAcquisitionAuctionCleanPrice' with 'IsBadBuy' to understand the pricing impact. Consider creating interaction terms or derived features based on these insights. If the data needs to be split, ensure that the investigation of these relationships is conducted on the training set to avoid data leakage.
	Simulated: True
	Score: avg score: 0.14587155963302753, simulated score: {'test_score': 0.15175644028103044, 'dev_score': 0.14587155963302753, 'score': 0.14587155963302753}, Visits: 2

		[Node 0-2-0]
		Plans: 
		4. Train a machine learning model using the preprocessed and feature-engineered training data. Before training, analyze the distribution of 'IsBadBuy' to understand the class imbalance. If necessary, apply appropriate techniques to handle the class imbalance during the data split and model training process.
		Simulated: False
		Score: avg score: 0, simulated score: {}, Visits: 0

		[Node 0-2-1]
		Plans: 
		4. Train a machine learning model using the preprocessed and feature-engineered training data. Before training, explore the correlation between 'VehicleAge' and 'IsBadBuy' to determine if older vehicles are more likely to be classified as 'IsBadBuy'. Ensure that this analysis is conducted on the training data split to avoid data leakage.
		Simulated: False
		Score: avg score: 0, simulated score: {}, Visits: 0

		[Node 0-2-2]
		Plans: 
		4. Train a machine learning model using the preprocessed and feature-engineered training data. Before training, investigate the relationship between 'MMRAcquisitionAuctionAveragePrice' and 'MMRAcquisitionAuctionCleanPrice' with 'IsBadBuy' to understand the pricing impact. Ensure that this analysis is conducted on the training set if the data is split.
		Simulated: False
		Score: avg score: 0.14587155963302753, simulated score: {'test_score': 0.15175644028103044, 'dev_score': 0.14587155963302753, 'score': 0.14587155963302753}, Visits: 1

		[Node 0-2-3]
		Plans: 
		4. Train a machine learning model using the preprocessed and feature-engineered training data. Before training, examine the frequency of 'Make' and 'Model' in the training data to identify which brands or models are more prone to being kicks. Ensure that this analysis is integrated with the data split process, if applicable, to maintain the distribution of 'Make' and 'Model' across the training and validation sets.
		Simulated: False
		Score: avg score: 0, simulated score: {}, Visits: 0

		[Node 0-2-4]
		Plans: 
		4. Train a machine learning model using the preprocessed and feature-engineered training data. Before training, visualize the distribution of 'VehOdo' to understand if high mileage vehicles are more likely to be problematic. Ensure that the data is split appropriately for training and validation, and integrate the insights from the 'VehOdo' distribution into the model training process.
		Simulated: False
		Score: avg score: 0, simulated score: {}, Visits: 0

	[Node 0-3]
	Plans: 
	3. Create new features that might improve the model's performance, such as interaction terms or domain-specific features. Additionally, examine the frequency of 'Make' and 'Model' to identify which brands or models are more prone to being kicks. Use these insights to create new features, such as a 'HighKickProneBrand' flag or a 'ModelKickFrequency' score, which can be integrated into the model. If the data is split, ensure that the frequency analysis is performed on the training set to avoid data leakage.
	Simulated: True
	Score: avg score: 0.09650833089037442, simulated score: {'test_score': 0.15175644028103044, 'dev_score': 0.14587155963302753, 'score': 0.14587155963302753}, Visits: 2

		[Node 0-3-0]
		Plans: 
		4. Train a machine learning model using the preprocessed and feature-engineered training data. Before training, analyze the distribution of 'IsBadBuy' to understand the class imbalance. If necessary, apply appropriate techniques to handle the class imbalance, such as oversampling, undersampling, or using class weights. Ensure that the data is split into training and validation sets, and perform the class imbalance analysis on the training set before applying any balancing techniques.
		Simulated: False
		Score: avg score: 0, simulated score: {}, Visits: 0

		[Node 0-3-1]
		Plans: 
		4. Train a machine learning model using the preprocessed and feature-engineered training data. Before training, explore the correlation between 'VehicleAge' and 'IsBadBuy' to determine if older vehicles are more likely to be classified as 'IsBadBuy'. Ensure that this exploration is conducted on the training set after any data splits have been performed.
		Simulated: False
		Score: avg score: 0, simulated score: {}, Visits: 0

		[Node 0-3-2]
		Plans: 
		4. Train a machine learning model using the preprocessed and feature-engineered training data. Before training, investigate the relationship between 'MMRAcquisitionAuctionAveragePrice' and 'MMRAcquisitionAuctionCleanPrice' with 'IsBadBuy' to understand the pricing impact. Ensure that this analysis is conducted on the training set if the data is split.
		Simulated: False
		Score: avg score: 0, simulated score: {}, Visits: 0

		[Node 0-3-3]
		Plans: 
		4. Train a machine learning model using the preprocessed and feature-engineered training data. Before training, examine the frequency of 'Make' and 'Model' in the training data to identify which brands or models are more prone to being kicks. Ensure that this analysis is performed on the training set if the data is split.
		Simulated: False
		Score: avg score: 0, simulated score: {}, Visits: 0

		[Node 0-3-4]
		Plans: 
		4. Before training the machine learning model using the preprocessed and feature-engineered training data, visualize the distribution of 'VehOdo' to understand if high mileage vehicles are more likely to be problematic. Ensure that this visualization is performed on the training set after splitting the data into training and validation sets.
		Simulated: True
		Score: avg score: 0.047145102147721316, simulated score: {'test_score': 0.06040992448759439, 'dev_score': 0.047145102147721316, 'score': 0.047145102147721316}, Visits: 1

	[Node 0-4]
	Plans: 
	3. Create new features that might improve the model's performance, such as interaction terms or domain-specific features. Additionally, visualize the distribution of 'VehOdo' to see if high mileage vehicles are more likely to be problematic. If the data is being split, ensure that the visualization is performed on the training set to avoid data leakage.
	Simulated: True
	Score: avg score: 0.11584490186692196, simulated score: {'test_score': 0.14014598540145987, 'dev_score': 0.11584490186692196, 'score': 0.11584490186692196}, Visits: 1

Generated 29 unique codes.
Best node: 0, score: {'test_score': 0.1530317613089509, 'dev_score': 0.14690841469084145, 'score': 0.14690841469084145}
Dev best node: 0, score: {'test_score': 0.1530317613089509, 'dev_score': 0.14690841469084145, 'score': 0.14690841469084145}
Grader score: 0.1530317613089509
\end{lstlisting}

\end{document}